\documentclass{article}

     \PassOptionsToPackage{numbers, compress}{natbib}

     \usepackage[preprint]{neurips_2019}

\usepackage[utf8]{inputenc} 
\usepackage[T1]{fontenc}    
\usepackage{hyperref}       
\usepackage{url}            
\usepackage{booktabs}       
\usepackage{amsfonts}       
\usepackage{nicefrac}       
\usepackage{microtype}      
\usepackage[pdftex]{graphicx}
\usepackage{xcolor} 
\usepackage[normalem]{ulem}
\useunder{\uline}{\ul}{}
\usepackage[ruled,linesnumbered]{algorithm2e}
\usepackage{tikz}
\usepackage{amsthm}
\usepackage{amsmath}
\usepackage{subcaption}
\usepackage{tikz-network}
\usepackage{pgfplots}
\usepackage{diagbox} 
\usepackage{caption}
\usepackage{siunitx}

\newcommand{\method}{FlowGN}

\theoremstyle{definition}
\newtheorem{definition}{Definition}[section]

\theoremstyle{plain}
\newtheorem{theorem}{Theorem}[section]

\title{Tracing the Propagation Path: A Flow Perspective of Representation Learning on Graphs}

\author{%
  Menghan~Wang \\
 Alibaba Group\\
   \texttt{xiangyu.wmh@alibaba-inc.com} \\
   \And
   Kun Zhang \\
   Department of Philosophy \\
   Carnegie Mellon University \\
   \texttt{kunz1@cmu.edu} \\
   \And
   Guli Lin \\
   Alibaba Group \\
   \texttt{guli.lingl@taobao.com} \\
   \And
   Keping Yang \\
   Alibaba Group \\
   \texttt{shaoyao@taobao.com} \\
   \And
   Luo si \\
   Alibaba Group \\
   \texttt{luo.si@alibaba-inc.com} \\
}

\begin{document}

\maketitle

\begin{abstract}
Graph Convolutional Networks (GCNs) have gained significant developments in representation learning on graphs. However, current GCNs suffer from two common challenges: 1) GCNs are only effective with shallow structures; stacking multiple GCN layers will lead to over-smoothing. 2) GCNs do not scale well with large, dense graphs due to the recursive neighborhood expansion. We generalize the propagation strategies of current GCNs as a \emph{``Sink$\to$Source''} mode, which seems to be an underlying cause of the two challenges. To address these issues intrinsically, in this paper, we study the information propagation mechanism in a \emph{``Source$\to$Sink''} mode. We introduce a new concept ``information flow path'' that explicitly defines where information originates and how it diffuses. Then a novel framework, namely Flow Graph Network (\method{}), is proposed to learn node representations. \method{} is computationally efficient and flexible in propagation strategies. Moreover, \method{} decouples the layer structure from the information propagation process, removing the interior constraint of applying deep structures in traditional GCNs. Further experiments on public datasets demonstrate the superiority of \method{} against state-of-the-art GCNs.
\end{abstract}

\section{Introduction}



Graph Convolutional Networks (GCNs) have become popular and powerful tools for representation learning on graphs. Motivated by CNNs, GCNs set a node’s one-hop graph neighborhood as its ``receptive field'' and perform convolution operation to aggregate feature information from the local neighborhood of the node.  By stacking multiple such convolutions information can be propagated across far reaches of a graph. However, two potential limitations prevent the further spread of GCNs. First, experiments \citep{li2018deeper,wang2018zero} show that GCNs are only effective with shallow structures (often no more than three layers) and the performance of GCNs drops dramatically when the number of layers increases, which indicates the fine-tuned GCNs are essentially shallow embedding. The shallow structure is still an open problem. Second, the recursive neighborhood expansion across layers poses time and memory challenges for training with large, dense graphs. A popular solution is to sample neighbors based on the `importance' of nodes to improve the scalability of GCNs \citep{hamilton2017inductive,chen2018fastgcn}.

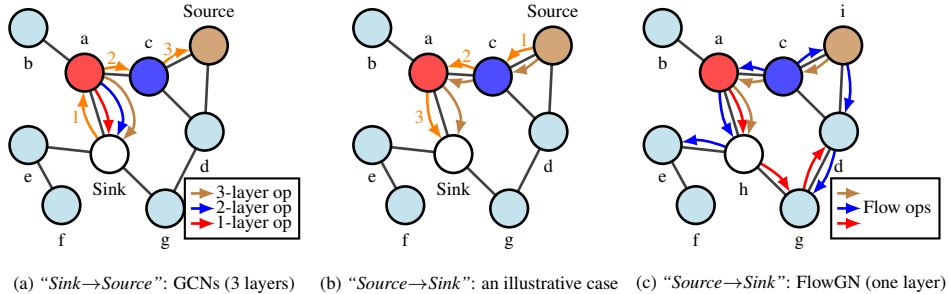
\begin{figure}
\centering
\begin{subfigure}{.3\textwidth}
\centering
\begin{tikzpicture}[scale=0.40]

\Vertex[x=2.868,y=5.518,size=0.5,color=red, opacity=0.7,position=above,label=a]{a};
\Vertex[x=1.000,y=7.000,size=0.5,opacity=0.7,position=below,label=b]{b};
\Vertex[x=5.006,y=5.387,size=0.5,color=blue,opacity=0.7,position=above,label=c]{c}
\Vertex[x=6.858,y=3.552,size=0.5,opacity=0.7,position=below,label=d]{d}
\Vertex[x=7.000,y=6.419,size=0.5,color=brown,opacity=0.7,position=above,label=Source]{e}
\Vertex[x=3.698,y=2.808,size=0.5,color=white,opacity=0.7,position=below,label=Sink]{f}
\Vertex[x=5.551,y=1.000,size=0.5, opacity=0.7,position=below,label=g]{g}
\Vertex[x=1,y=3.108,size=0.5,opacity=0.7,position=below,label=e]{h}
\Vertex[x=2.1,y=1.108,size=0.5,opacity=0.7,position=below,label=f]{i}


\Edge[,lw=1.0,Direct=false](a)(b)
\Edge[,lw=1.0,Direct=false](a)(c)
\Edge[,lw=1.0,Direct=false](c)(d)
\Edge[,lw=1.0,Direct=false](d)(e)
\Edge[,lw=1.0,Direct=false](e)(c)
\Edge[,lw=1.0,Direct=false](f)(h)
\Edge[,lw=1.0,Direct=false](i)(h)
\Edge[,lw=1.0,Direct=false](f)(g)
\Edge[,lw=1.0,Direct=false](d)(g)
\Edge[,lw=1.0,Direct=false](f)(a)

\Edge[,lw=1.0,bend=15,color=red,Direct,](a)(f)
\Edge[,lw=1.0 ,bend=40.,color=blue,Direct](a)(f)
\Edge[,lw=1.0 ,bend=60.,color=brown,Direct](a)(f)


\Edge[,lw=1.0,bend=20,color=orange,Direct=true,label = 1, position= left](f)(a)
\Edge[,lw=1.0 ,bend=15.,color=orange,Direct=true,label = 2, position= above, distance=0.3](a)(c) \Edge[,lw=1.0 ,bend=15.,color=orange,Direct=true,label = 3, position= above, distance=0.2 ](c)(e)



\def\x{6.2}
\draw[black, thick] (\x,0) rectangle (10,2);
\Edge[,lw=1.0,color=red,Direct](\x+.2,0.5)(\x+1 ,0.5)
\Edge[,lw=1.0,color=blue,Direct](\x+.2,1)(\x+1. ,1)
\Edge[,lw=1.0,color=brown,Direct](\x+.2,1.5)(\x+1. ,1.5)
\node at (\x+2.3 ,0.5) {\scriptsize 1-layer op};
\node at (\x+2.3 ,1) {\scriptsize 2-layer op};
\node at (\x+2.3 ,1.5) {\scriptsize 3-layer op};
 


\end{tikzpicture}
\captionsetup{font=scriptsize}
\subcaption{\emph{``Sink$\to$Source''}: GCNs (3 layers)}\label{graph_gcns}
\end{subfigure}%
\begin{subfigure}{.3\textwidth}
\centering
\begin{tikzpicture}[scale=0.40]
\Vertex[x=2.868,y=5.518,size=0.5,color=red, opacity=0.7,position=above,label=a]{a};
\Vertex[x=1.000,y=7.000,size=0.5,opacity=0.7,position=below,label=b]{b};
\Vertex[x=5.006,y=5.387,size=0.5,color=blue,opacity=0.7,position=above,label=c]{c}
\Vertex[x=6.858,y=3.552,size=0.5,opacity=0.7,position=below,label=d]{d}
\Vertex[x=7.000,y=6.419,size=0.5,color=brown,opacity=0.7,position=above,label=Source]{e}
\Vertex[x=3.698,y=2.808,size=0.5,color=white,opacity=0.7,position=below,label=Sink]{f}
\Vertex[x=5.551,y=1.000,size=0.5, opacity=0.7,position=below,label=g]{g}
\Vertex[x=1,y=3.108,size=0.5,opacity=0.7,position=below,label=e]{h}
\Vertex[x=2.1,y=1.108,size=0.5,opacity=0.7,position=below,label=f]{i}


\Edge[,lw=1.0,Direct=false](a)(b)
\Edge[,lw=1.0,Direct=false](a)(c)
\Edge[,lw=1.0,Direct=false](c)(d)
\Edge[,lw=1.0,Direct=false](d)(e)
\Edge[,lw=1.0,Direct=false](e)(c)
\Edge[,lw=1.0,Direct=false](f)(h)
\Edge[,lw=1.0,Direct=false](i)(h)
\Edge[,lw=1.0,Direct=false](f)(g)
\Edge[,lw=1.0,Direct=false](d)(g)
\Edge[,lw=1.0,Direct=false](f)(a)

\Edge[,lw=1.0,bend= -20,color=orange,Direct=true,label = 3, position= left](a)(f)
\Edge[,lw=1.0 ,bend=-20.,color=orange,Direct=true,label = 2, position= above, distance=0.3](c)(a)
\Edge[,lw=1.0 ,bend=-20.,color=orange,Direct=true,label = 1, position= above, distance=0.3](e)(c)



\Edge[,lw=1.0,bend= 15,color=brown,Direct](e)(c)
\Edge[,lw=1.0,bend= 15,color=brown,Direct](c)(a)
\Edge[,lw=1.0,bend= 30,color=brown,Direct](a)(f)


\end{tikzpicture}
\captionsetup{font=scriptsize}
\subcaption{\emph{``Source$\to$Sink''}: an illustrative case }
\label{sketch}
\end{subfigure}
\begin{subfigure}{.3\textwidth}
\centering
\begin{tikzpicture}[scale=0.40]
\Vertex[x=2.868,y=5.518,size=0.5,color=red, opacity=0.7,position=above,label=a]{a};
\Vertex[x=1.000,y=7.000,size=0.5,opacity=0.7,position=below,label=b]{b};
\Vertex[x=5.006,y=5.387,size=0.5,color=blue,opacity=0.7,position=above,label=c]{c}
\Vertex[x=6.858,y=3.552,size=0.5,opacity=0.7,position=below,label=d]{d}
\Vertex[x=7.000,y=6.419,size=0.5,color=brown,opacity=0.7,position=above,label=i]{e}
\Vertex[x=3.698,y=2.808,size=0.5,color=white,opacity=0.7,position=below,label=h]{f}
\Vertex[x=5.551,y=1.000,size=0.5, opacity=0.7,position=below,label=g]{g}
\Vertex[x=1,y=3.108,size=0.5,opacity=0.7,position=below,label=e]{h}
\Vertex[x=2.1,y=1.108,size=0.5,opacity=0.7,position=below,label=f]{i}


\Edge[,lw=1.0,Direct=false](a)(b)
\Edge[,lw=1.0,Direct=false](a)(c)
\Edge[,lw=1.0,Direct=false](c)(d)
\Edge[,lw=1.0,Direct=false](d)(e)
\Edge[,lw=1.0,Direct=false](e)(c)
\Edge[,lw=1.0,Direct=false](f)(h)
\Edge[,lw=1.0,Direct=false](i)(h)
\Edge[,lw=1.0,Direct=false](f)(g)
\Edge[,lw=1.0,Direct=false](d)(g)
\Edge[,lw=1.0,Direct=false](f)(a)


\Edge[,lw=1.0,bend=15,color= red,Direct](a)(f)
\Edge[,lw=1.0,bend=15,color= red,Direct](f)(g)
\Edge[,lw=1.0,bend=15,color= red,Direct](g)(d)

\Edge[,lw=1.0,bend=-15,color=blue,Direct](c)(a)
\Edge[,lw=1.0,bend=-15,color=blue,Direct](a)(f)
\Edge[,lw=1.0,bend=-15,color=blue,Direct](f)(h)

\Edge[,lw=1.0,bend= 15,color=blue,Direct](c)(e)
\Edge[,lw=1.0,bend= 15,color=blue,Direct](e)(d)
\Edge[,lw=1.0,bend =15,color=blue,Direct](d)(g)

\Edge[,lw=1.0,bend= 15,color=brown,Direct](e)(c)
\Edge[,lw=1.0,bend= 15,color=brown,Direct](c)(a)
\Edge[,lw=1.0,bend= 30,color=brown,Direct](a)(f)

\def\x{6.6}
\draw[black, thick] (\x,0) rectangle (10.2,2);
\Edge[,lw=1.0,color=red,Direct](\x+.2,0.5)(\x+1 ,0.5)
\Edge[,lw=1.0,color=blue,Direct](\x+.2,1)(\x+1,1)
\Edge[,lw=1.0,color=brown,Direct](\x+.2,1.5)(\x+1,1.5)
\node at (\x+2.2,1.) {\scriptsize Flow ops};

\end{tikzpicture}
\captionsetup{font=scriptsize}
\subcaption{\emph{``Source$\to$Sink''}: FlowGN (one layer)}
\label{flowgn_example}
\end{subfigure}
\captionsetup{font=footnotesize}
\caption{(a) and (b) are two examples of information propagation between the \emph{Source} node and the \emph{Sink} node. The \textcolor{orange}{orange} arrows indicate effective computation paths with order, and the other arrows (\textcolor{brown}{brown}, \textcolor{red}{red}, and \textcolor{blue}{blue}) indicate the concrete computation operations and their color indicates the \emph{Source} node whose information is expected to be transmitted in the corresponding operation. GCNs (left) recursively request node $a$ for information, and finally gets the information of the \emph{Source} node at the $3^{rd}$ layer (we omit some arrows for simplification). The effective computation path is $\{Sink \to a \to c \to Source\}$. The illustrative case (middle) passes information from the \emph{Source} node to the \emph{Sink} node directly with three hops (The brown arrows). Its effective computation path is $\{Source \to c \to a \to Sink\}$. (c) is an example of FlowGN where the path length is $3$.}
\label{example}
\end{figure}

The key to aforementioned challenges lies in how we understand and model the information propagation mechanism. \citet{xu2018representation} showed that the neighborhood aggregation schemes for a $K$-layer GCN is analogous to the spread of a $K$-step random walker with some mild assumptions. At a micro level, the representation learning process of one node from another node in GCN can be viewed as computation paths that connect the two nodes. Figure \ref{graph_gcns} shows an example that a \emph{Sink} node recursively learn information from a \emph{Source} node through one computation path. We generalize this kind of information propagations as a \emph{``Sink$\to$Source''} mode, i.e., the learning process starts from the \emph{Sink} node, and seeks source nodes recursively to request information from them. Unfortunately, the \emph{``Sink$\to$Source''} mode has two unfavorable characteristics: 1) Potential \emph{Source} nodes are not known to \emph{Sink} nodes in advance. A \emph{Sink} node has to recursively seek \emph{Source} nodes so it needs $K$ neighborhood aggregations in order to learn from the \emph{Source} nodes that are $K$ hops away. 2) The pattern of information diffusion is implicit and difficult to evaluate. It is hard to estimate how much information of a \emph{Source} node is lost (or washed out by aggregation operations) along the computation path before reaching the \emph{Sink} node. These two characteristics restrict the capacity of GCNs together: in order to extend propagation ranges, GCNs have to stack more layers and run numerous aggregation operations to transmit information, which, in turn, leads to more information loss and duplicated computation. The \emph{``Sink$\to$Source''} mode may be one of the underlying causes of the shallow embedding and the scalability problems in GCNs.

Alternatively, one can model the information propagation process in a \emph{``Source$\to$Sink''} mode: starting from \emph{Source} nodes and seeking \emph{Sink} nodes iteratively in graphs (Figure \ref{sketch} shows an example). We can explicitly define where information originates and how it diffuses, which could circumvent the limitations of the \emph{``Sink$\to$Source''} mode. As we know all the source nodes in advance, we can model information propagation with variable range in one layer; we do not have to perform propagations recursively. Moreover, a proper diffusion mechanism design will reduce unnecessary computation. For instance, we can assume information transmission in a computation path is only dependent on the nodes in the path, removing all the neighborhood aggregations of the nodes along the path. We argue that the \emph{``Source$\to$Sink''} mode could be a plausible direction in representation learning on graphs.

Particularly, in this paper, we study the information propagation mechanism of GCNs in a \emph{``Source$\to$Sink''} mode. We introduce a new concept, ``information flow path", that explicitly defines where information originates and how it diffuses. Then a novel framework, namely Flow Graph Network (\method{}), is proposed to learn node representations on graphs. \method{} is flexible in propagation strategies and computationally efficient. Moreover, \method{} decouples the layer structure from the information propagation process, removing the interior constraint of applying deep structures in current GCNs. Further experiments on public datasets reveal the superiority of \method{}.



\section{Related Work}
In this section, we first introduce some related work on GCNs and random walk-based methods with respect to representation learning. Then we describe some background materials on network flow and centrality measures, which are related to our proposed method.  

The core idea behind GCNs is to learn how to iteratively aggregate feature information from local graph neighborhoods. Advances in this direction are often categorized as spectral approaches and non-spectral approaches. Some spectral approaches \citep{BrunaZSL13,kipf2016semi} define convolution operation in the Fourier domain by computing the eigendecomposition of the graph Laplacian. Non-spectral approaches define convolutions directly on graphs and operate on spatially close neighbors. GraphSAGE \citep{hamilton2017inductive} generates embeddings by sampling and aggregating features from a node’s local neighborhood. The FastGCN model \citep{chen2018fastgcn} treats graph convolutions as integral transforms of embedding functions. Different from GraphSAGE, FastGCN samples vertices rather than neighbors for aggregation. \citet{huang2018adaptive} proposes a layer-wise sampling method to capture between-layer correlations. 
The GAT model \cite{velickovic2017graph} uses a self-attention mechanism to learn content aggregation weights between neighbors and the current node for graph representation learning. 
Another line of related work of representation learning is random walk methods (e.g., DeepWalk \citep{perozzi2014deepwalk}, LINE \citep{tang2015line}, and Node2Vec \citep{grover2016node2vec}). They treat nodes as words and the generated random walks on graphs as sentences, and then apply SkipGram model on them. Random walk methods are effective and easy to implement; they have been widely applied in network applications to learn node representations. However, these methods also lead to potential drawbacks: 1) embeddings of nodes are learned independently and parameters are not shared; it is computationally inefficient since the number of parameters grows as $O(|\mathcal{V}|)$. 2) Node attributes are not utilized, which contain rich information w.r.t. node representations. Recently, \citet{klicpera2018predict} combines neural network with personalized pagerank, which alleviates the above two issues and achieves improvements on representation learning.

Graphs are widely studied in network theory \citep{wasserman1994social} for various problems and the corresponding findings may provide insight to representation learning. Centrality \citep{freeman1978centrality}, a fundamental concept in network theory, identifies the importance of nodes within a graph. Centrality measures take into account how a node interacts and communicates with the rest of the graph and have proved of value in understanding of the role played by the nodes in the graph.  Intuitively, centrality assesses a node’s involvement in the structure of a graph and thus can be utilized for neighborhood aggregation (node importance) in GCNs. Particularly, \citet{borgatti2005centrality} viewed centrality as a node-level outcome of implicit models of flow processes and showed centrality choice should match the flow characteristics of networks, which gave us the initial inspiration that a proper flow process may efficiently model the information propagation for representation learning on graphs.

\section{FlowGN}
In this section, we introduce our Flow Graph Network (\method{}) framework in detail. 
Assume we have a graph  $\mathcal{G(V,E)}$ with node feature $X_{v}$ for $v \in \mathcal{V}$. \method{} is a $K$-layer model and its output is feature vector $z_{v}$ for all $v \in \mathcal{V}$. The quality of learned representations $z_{v}$ is evaluated by downstream tasks, such as link prediction and node classification. A graphical example of \method{} is shown in Figure \ref{flowgn_example}. In the remaining parts of this section, we first introduce the information flow algorithm (Section \ref{3.1}). Then we describe the \method{} embedding generation (i.e., forward propagation) algorithm and its inference (Section \ref{3.2}). In Section \ref{3.3} we further discuss the characteristics of \method{}.



\subsection{Information Flow Algorithm}\label{3.1}
Below we give the definitions of the information flow path and the information propagation mechanism that explicitly describe the information diffusion process in a \emph{``Source$\to$Sink''} mode. 
 
 
\begin{definition}[Information Flow Path]
Given a graph  $\mathcal{G(V,E)}$, an information flow path is defined as $P=(s\xrightarrow{e_{0}}v_{1}\xrightarrow{e_{1}}v_{2}\xrightarrow{e_{2}}...v_{n}\xrightarrow{e_{n}}t)$, where $s$ is the source node, $\{v_{i}\}$ is the set of intermediate nodes, $t$ is the sink node, and $\{e_{i}\}$ is the set of edges that connect nodes along the path. The path length $l$ is set as the number of nodes in $P$. 

\end{definition}

\begin{definition}[Information Propagation mechanism]
An information propagation mechanism is a meta template defined on an information flow path and is denoted with three functions: $GenerateFlow(v)$, $TransmitFlow(v, flow)$, and $ConserveFlow(v, flow)$. The procedure of propagation is as follows: the source node $s$ first produces information flow via $GenerateFlow(s)$ and then transmits the information flow along the path. For every intermediate node $\{v_{i}\}$ along the path, it receives the information flow from the upstream node and calls $ConserveFlow(v, flow)$ and $TransmitFlow(v, flow)$ separately. These two functions indicate the node $\{v_{i}\}$ conserves information and transmits flow to the downstream node. Finally, the flow ends at the sink node $t$ after calling $ConserveFlow(t, flow)$. A detailed implementation of the propagation procedure is shown in Algorithm \ref{algo1}.  
\end{definition}

\begin{algorithm}[tb]
\small
\SetAlgoLined
\KwIn{Graph $\mathcal{G(V,E)}$; path iteration $r$; path length $l$; return $p$; in-out $q$  }
\KwOut{Information flow paths $\mathcal{P}_{m}$}
Initialization: set $\mathcal{P}_{m}$ to empty\;
 \For{$iter=1...r$}
 {
 	 \For{$v \in \mathcal{V}$}
 	{	
		$RestartNum$ = IMPORTANCE($\mathcal{G},v$) \tcc*{Centrality measures}
 		 \For{$i=1...RestartNum$}
 		{	
 			$path$ = node2vecWalk($l, p, q$)  \tcc*{  \citet{grover2016node2vec} }
			append $path$ to $\mathcal{P}_{m}$
 		}
 	}
 }
 
\caption{Path generation algorithm (PATHGEN)}
\label{pathgeneration}
\end{algorithm}



\begin{algorithm}[tbh]
\small
\SetAlgoLined
\KwIn{hidden features $\{h_{v}, \forall v \in \mathcal{V}\}$;information flow paths $\{\mathcal{P}_{m},\forall m \in \mathcal{M} \}$}
\KwOut{Vector representations $h_{N(v)}$ for all $v\in V$}
Initialization: set $h_{C(v)}$ to empty for all $v\in V$\tcc*{$h_{C(v)}$ is a set that stores the conserved flows}
 \For{$m=1...\mathcal{M}$} 
 {  
    $s, \{v_{n}\},t$ = $\mathcal{P}_{i}$\; 
    currentflow = $GenerateFlow(s)$  \tcc*{current flow is the same size as $h_{s}$}
    
    \For{$i=1...n$}
    {   
         $h_{C(v_{i})} \leftarrow ConserveFlow(v_{i}, currentflow)$\;
        currentflow = $TransmitFlow(v_{i}, currentflow)$\;
        
    }
    
    $h_{C(t)} \leftarrow ConserveFlow(t, currentflow)$ \;
 }
 \For{$v \in \mathcal{V}$}
 {
 $h_{N(v)}$ = AGGREGATE$(h_{C(v)})$  \tcc*{neighborhood aggregation}
 }
\caption{Information Propagation (INFOPROPAGATE) }
\label{algo1}
\end{algorithm}

The formula of our information flow algorithm is inspired by the classical network flow problem \citep{ahuja1988network}. Analogously, we treat feature transmission as flows and the $TransmitFlow(v, flow)$ function is used to control flows (similar to the `capacity' concept in network flow problem). One difference is that \method{} allows intermediate nodes to conserve flows ($ConserveFlow(v, flow)$) in order to encourage reuse of flow paths.


\textbf{Flow path generation.} The first step of path generation is to choose the source node. In a flow path only the source node's information is propagated and our goal is to learn node representations with structural information, so we generate information flow paths starting from every node in the graph. Although a na\"ive approach is to run $r$ paths for each start node, a more reasonable setting is to let ``important'' nodes generate more flow paths in a graph as they are more influential in the graph. In graph theory, centrality is a measure that identifies the importance of vertices within a graph. Here we choose degree centrality (i.e., the number of edges a node has) to denote the importance score of nodes. After we choose a source node, the rest part of a flow path can be generated with a variety of random walk algorithms, which are computationally efficient in terms of both space and time requirements. We choose node2vec \citep{grover2016node2vec} as a base model to generate information flow paths because it flexibly interpolates between depth-first search (DFS) and breadth-first search (BFS) strategies. We fix path length to $l$ for model simplification. The pseudocode of flow path generation is given in Algorithm \ref{pathgeneration}.


\subsection{Embedding Learning}\label{3.2}
We describe the embedding learning of FlowGN in Algorithm \ref{embedding generation algorithm}. Specifically, in $k^{th}$ layer, FlowGN first generates information flow paths, and run Algorithm \ref{algo1} to propagate information. FlowGN then concatenates the node’s representation in $(k-1)^{th}$ layer, $h_{v}^{k-1}$, with the aggregated neighborhood vector, $h^{k}_{v}$ , and this concatenated vector is fed through a fully connected layer with nonlinear activation function $\sigma$. The output $h_{v}^{k}$ is used as the hidden features of node $v$ in the next layer. 

\begin{algorithm}[tbh]
\small
\SetAlgoLined
\KwIn{Graph $\mathcal{G(V,E)}$; input features $\{x_{v}, \forall v \in V\}$;non-linearity $\sigma$; depth $K$; different aggregator functions; number of information flow paths $\mathcal{M}$ }
\KwOut{Vector representations $z_{v}$ for all $v\in V$}
$h_{v}^{0} \leftarrow x_{v}, \forall v \in \mathcal{V} \;$
 
 \For{$k=1...K$}
 {
 $\mathcal{P}_{m}$ = PATHGEN($\mathcal{G},\mathcal{M}$)  \tcc*{Generate flow paths Algorithm \ref{pathgeneration}}
 
 $h^{k}_{N(v)} \leftarrow$ INFOPROPAGATE($\{h_{v}^{k-1}, \forall v \in \mathcal{V} \}$, $\{\mathcal{P}_{m},\forall m \in \mathcal{M} \}$) \tcc*{Algorithm \ref{algo1}}
 $h^{k}_{v} \leftarrow \sigma\big(W^{k} \cdot$ CONCAT$( h^{k-1}_{v},h^{k}_{N(v)})\big)$, $\forall v \in \mathcal{V}$
 }
 $z_{v} \leftarrow h^{K}_{v}, \forall v \in \mathcal{V}$
\caption{FlowGN embedding generation algorithm}\label{embedding generation algorithm}
\end{algorithm}

\textbf{Neighborhood aggregation}. In each layer of \method{}, every node only needs one neighborhood aggregation and the neighbors are not limited to adjacent nodes. Due to the randomness of flow paths, the neighbors are not ordered. An ideal aggregator function would be symmetric, i.e., the aggregation results should be invariant to the order of the neighbors. In \method{}, we choose the mean aggregator function, where we simply take the element-wise mean of the vectors in $h_{C(v)}$.

\textbf{Inference}. FlowGN can be trained with unsupervised or supervised loss functions, depending on the specific tasks, and the parameters can be learned using standard stochastic gradient descent and backpropagation techniques. Similar to GraphSAGE \citep{hamilton2017inductive}, FlowGN learned a set of aggregator functions, which could be further utilized to test unseen nodes. In experiments, we apply a mini-batch learning setting to speed up the model training. 

\subsection{Comparison with Related Work}\label{3.3}
\textbf{Weighted neighborhood aggregation and information flow.}  Recent studies increased the capacity of GCNs via assigning different importance (weights) to nodes of a same neighborhood. They either use self-attention layers or develop different kinds of neighborhood sampling methods. FlowGN does the same thing but with a different approach; it is implicitly implemented in the information propagation process. Specifically,  a \emph{sink} node may receive many flows from the same \emph{source} node through flow paths, and the flow counts can be viewed as unnormalized weights. Compared with other counterparts, FlowGN requires no additional model modification or computation.

 
\textbf{Layer structure and information propagation.} The propagation range of current GCNs is closely tied to the number of convolutional layers, i.e., in each layer nodes aggregate features from their 1-hop neighbors so the layer number explicitly determines the propagation bounds for all nodes. Once the neural structure is determined, the propagation range for all nodes are fixed. In FlowGN, the propagation is determined by the path length $l$. Potential benefits are two-fold. On one hand, it's more flexible to determine the propagation range. On the other hand, we can explore deep structures by stacking layers without any constraint from the information propagation mechanism.



\textbf{Homogeneous or heterogeneous.} Most existing GCNs implicitly assume a homogeneous graph setting, but in real scenarios many graphs are heterogeneous, where multiple types of nodes interact via different types of relationships. Different types of nodes have different feature spaces, which brings two challenges to information propagation: 1) how to define neighbors for a specific node and 2) how to aggregate information among different feature spaces. Unlike GCNs that need complicated model modifications, \method{}  can seamlessly be incorporated with the `standard' approach in heterogeneous settings; we assume $\mathcal{G(V,E)}$ is associated with a node type mapping function $\phi:\mathcal{V} \rightarrow \mathcal{A}$ and a link type mapping function $\varphi:\mathcal{E} \rightarrow \mathcal{R}$ ($\mathcal{A}$ and $\mathcal{R}$ denote the sets of object types and link types). Then a straightforward approach could be to define the information flow path in a Meta-path manner \cite{sun2011pathsim}, and assign a type-specific transformation matrix in $ConserveFlow(v, flow)$.

\textbf{Relation to random walk methods.} As opposed to random walk based methods, \method{} utilizes node attributes and the parameters of \method{} are shared among nodes to make the model efficient and regularized. With a multi-layer structure, the learned node representations of \method{} are not limited in shallow embedding. At the same time, \method{} inherits the merit of random walk methods during its information flow path generation, which is computationally efficient in terms of both time and memory requirements.

\section{Therotical Analysis }
In this section we probe the relations between GCNs (\emph{``Sink$\to$Source''}) and \method{} (\emph{``Source$\to$Sink''}) in order to provide insight into how \method{} can learn about graph structure. \citet{xu2018representation} showed that the process of information propagation in GCNs is analogous to the spread of a random walker.  Following their ideas and settings (randomization assumption of ReLU \citep{Kawaguchi16} and Influence distribution \citep{KohL17}), we can draw connections between GCNs and \method{} with the property of random walks, which comes to the following theorem:
\begin{theorem}\label{t1}
Given a grid graph  $\mathcal{G(V,E)}$, and a k-layer GCN with averaging as its neighborhood aggregation scheme, there exists an equivalent version of \method{} that has the same influence distribution for every node $x \in \mathcal{V}$. 
\end{theorem}

Theorem \ref{t1} indicates \emph{``Sink$\to$Source''} and \emph{``Source$\to$Sink''} are symmetry in grid graphs with some mild assumptions. The full proof of theorem \ref{t1} is in the Appendix. From the proof we can also observe \method{} acquire less computation than GCNs along the computation paths; \method{} needs one weight matrix in the expected influence distribution while GCNs need a multiplication of $K$ weight matrices. In reality, most graphs are often non-grid and thus the two modes are asymmetry. Consider a task of infection risk prediction in an epidemic network, where some infectious disease spreads from infected individuals to healthy individuals along the network structure. We argue that the \emph{``Source$\to$Sink''} mode should be chosen in this case as it represents the true mechanism of information propagation (disease diffusion); improper choice may lead to inductive biases. We believe that our work would shed new light on the underlying mechanism of information propagation in graphs for representation learning.

\section{Experiments}
We implement a simple yet effective version of information propagation mechanism that enables \method{} to learn better representations with less time complexity. Specifically, we set $GenerateFlow(s)=h_{s}$, $TransmitFlow(v_{i}, flow)=flow$, and $ConserveFlow(v_{i},flow)= flow$. This setting assumes that information transmits along flow paths with no information loss. We test the performance of \method{} on four benchmark datasets: \emph{Cora}, \emph{Citeseer}, \emph{Pubmed} \citep{sen2008collective}, and \emph{Coauthor} \citep{shchur2018pitfalls}. The first three datasets are citation datasets, in which every node represents a paper and the edges represent citations between them. The last dataset is about academic coauthor relationships, where nodes indicate authors and edges indicate coauthor relations. Our goal is to classify academic papers into different subjects using the citation datasets. We use prediction accuracy as our main metric. Following the experiment setup in \cite{chen2018fastgcn}, we adjusted the original training/validation/test split of \emph{Cora}, \emph{Citeseer}, and \emph{Pubmed} to align with the supervised learning scenario. Specifically, all labels of the training examples are used for training.The split of \emph{Coauthor} is coherent with that of other datasets. The statistics of the experimental datasets are shown in Table \ref{table:citation-dataset}.

\begin{table}[tbh]
\small
\centering
\begin{tabular}{l|cccccc}
\toprule
Dataset & \# Nodes & \# Edges & \# Classes & \# Features &\# Avg. SP & Train/Vali/Test Nodes \\
\midrule
Cora & $2,708$ & $5,429$ &7 & $1,433$& $6.31$ & $1,208/500/1,000$\\
Citeseer & $3,327$ & $4,732$ &6& $3,703$&$9.33$ &  $1,812/500/1,000$\\
Pubmed & $19,717$ & $44,338$ &3& $500$&$6.34$ &  $18,217/500/1,000$\\
CoAuthor & $18,333$ & $81,894$  & $15$ & $6,805$&$5.43$ & $16,833/500/1,000$\\
\bottomrule
\end{tabular}
\caption{Dataset statistics of experimental datasets. Avg. SP means average shortest path length.}
\label{table:citation-dataset}
\end{table}

\textbf{Baselines.} We compare FlowGN against  node2vec \citep{grover2016node2vec}, GCN \citep{kipf2016semi}, GAT \citep{velickovic2017graph}, GraphSAGE \citep{hamilton2017inductive}, FastGCN \citep{chen2018fastgcn}, and SGC \citep{sgc} using the publicly released implementations. All baseline models contain two layers if not specified. Due to space limit, we omit part of experimental settings and results of \emph{Cora} below and please refer to the Appendix for more details.


\subsection{Performance}  
We report the performance of our method and baseline models in Table \ref{table:baselines}. As shown in the results, we can see FlowGN outperforms other baseline models in average. Node2vec performs worst among baselines because the embeddings are learned independently and node features are not utilized. SGC is competitive but not stable; it achieves best performance in \emph{Cora} but performs poorly in \emph{Pubmed}. FastGCN samples nodes via importance and suffers from a high variance problem, which may introduce biases and influence the prediction accuracy. Note that all baselines except Node2vec actually propagate information in a \emph{``Sink$\to$Source''} mode. The superiority of \method{} in prediction accuracy demonstrates the effectiveness of the \emph{``Source$\to$Sink''} mode.  

\begin{table}[htb!]
\small
\centering
         \begin{tabular}{l|c|c|c|c}
        \toprule
         Model & Cora & Citeseer & Pubmed &CoAuthor CS \\ 
        \midrule
       node2vec  & $19.73\pm{0.95}$ & $18.02\pm{1.12}$ & $33.91\pm{1.32}$&$11.50\pm{1.20}$ \\
        GCN(original) & $86.10\pm{0.40}$ & $\emph{77.60}\pm \emph{0.40}$ & $86.60\pm{0.30}$ & $90.90\pm{0.30}$ \\
        GAT & $82.39\pm{0.71}$ & $75.49\pm{0.65}$ &  $84.30\pm{0.33}$ &$90.72\pm{0.27}$ \\
        FastGCN & $85.90\pm{0.30}$ & $76.10\pm{0.60}$ & $87.90\pm{0.30}$ & $92.90\pm{0.40}$ \\
        SGC & $ \textbf{87.00}\pm \textbf{0.00}$ &  $76.90 \pm{0.00}$ & $82.90\pm{0.00} $ & $91.40\pm{0.00}$ \\
        GraphSAGE-mean & $82.20\pm{0.47} $ & $68.56\pm{0.91} $  & $\emph{88.80}\pm \emph{0.56}$ &$\emph{93.01} \pm \emph{0.45}$ \\
        {\method{}  } & $\emph{86.32}\pm \emph{0.30}$  & $\textbf{79.76}\pm \textbf{0.42}$ & $\textbf{89.41}\pm \textbf{0.27}$ &$\textbf{94.85} \pm \textbf{0.11}$ \\
         \bottomrule
        \end{tabular}
        \caption{Test accuracy (\%) on four datasets (Results are averaged over 10 runs).}
\label{table:baselines}        
\end{table}

\subsection{ Analysis of Flow Path} \label{ana_flow_path}
\begin{figure}[thb]
\centering
\begin{subfigure}{.3\textwidth}
\includegraphics[height=3.7cm]{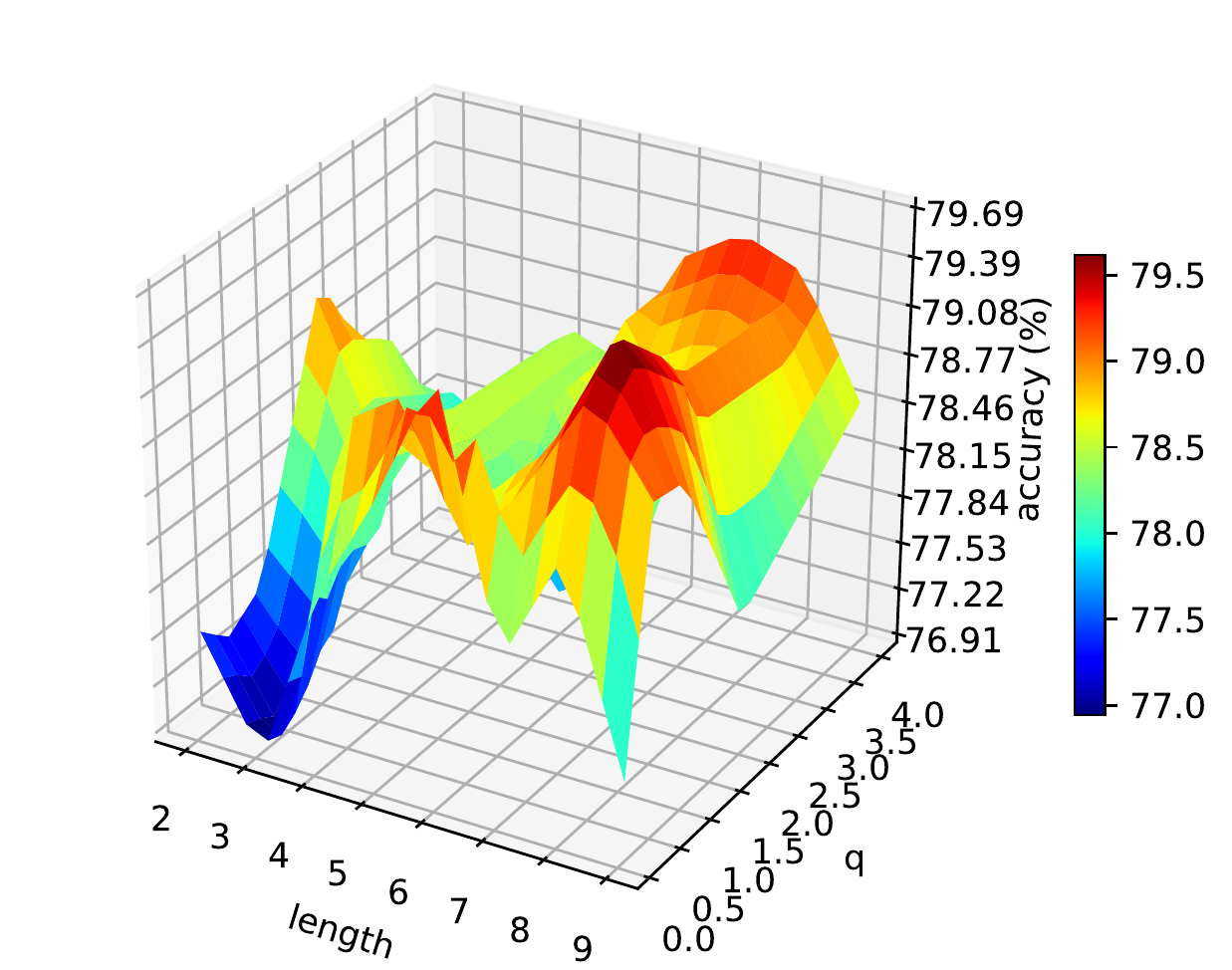}
 \caption{Citeseer}
\end{subfigure}
\quad 
\begin{subfigure}{.3\textwidth}
\includegraphics[height=3.7cm]{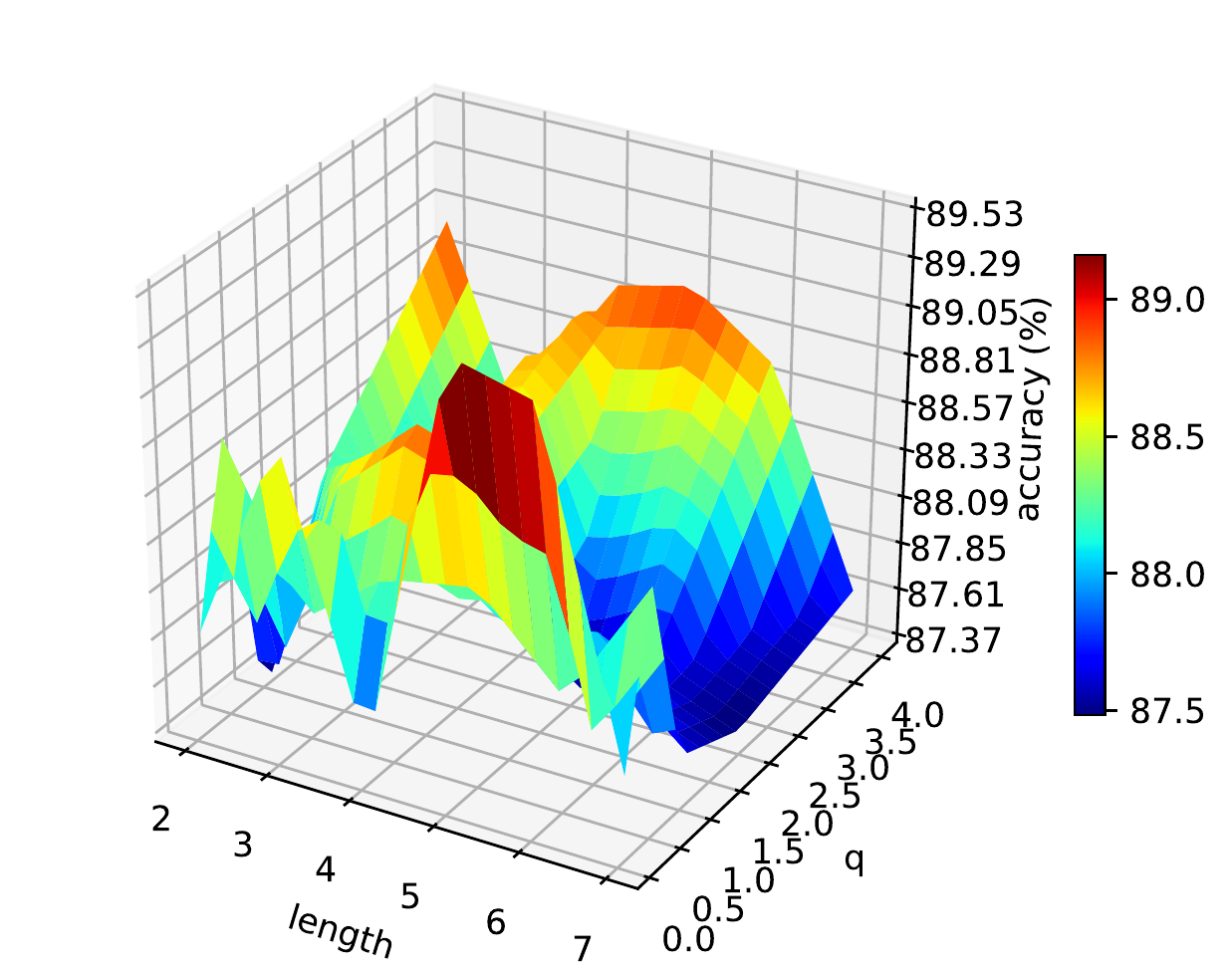}
 \caption{Pubmed}
\end{subfigure}
\quad 
\begin{subfigure}{.3\textwidth}
 \includegraphics[height=3.7cm]{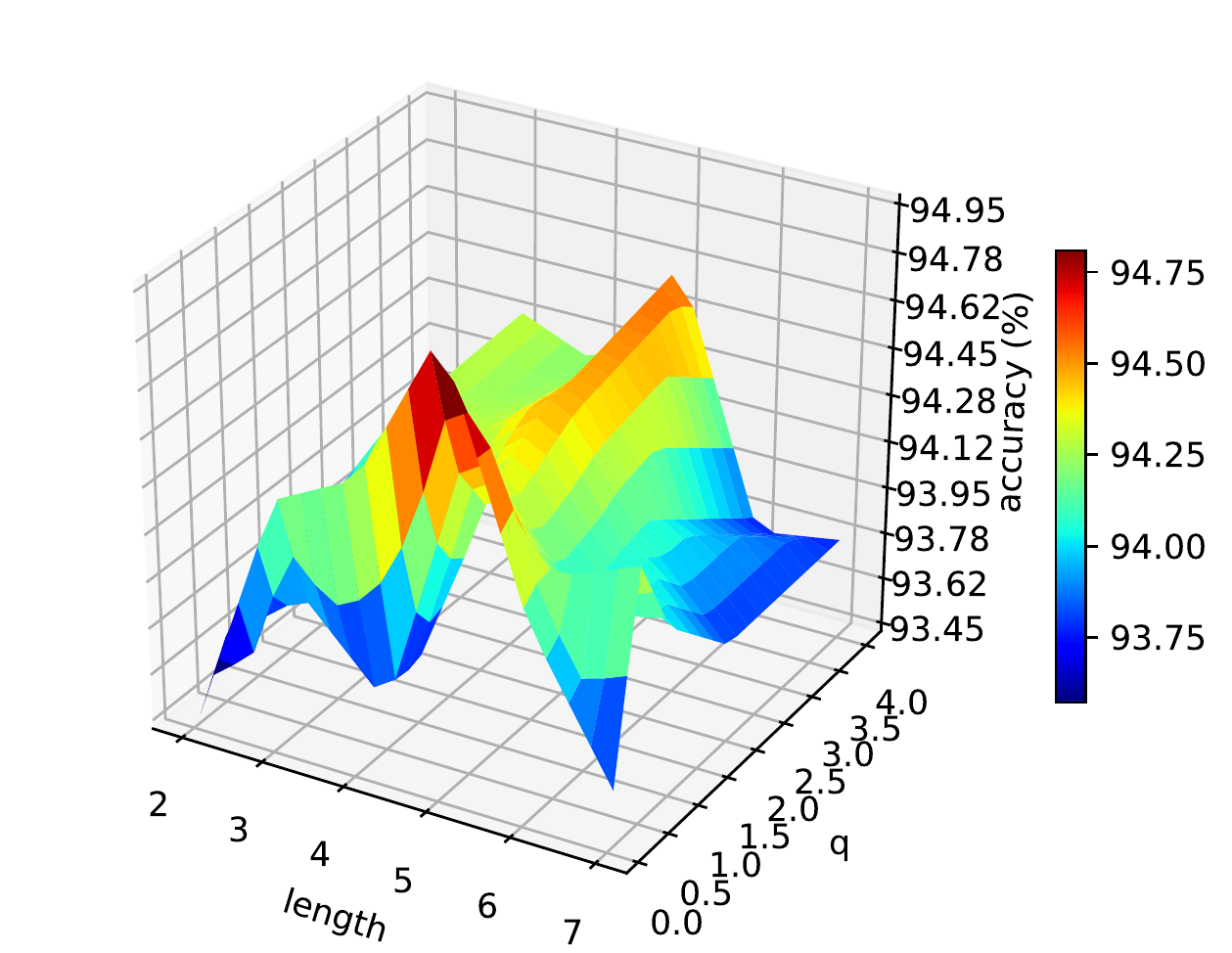}
 \caption{Coauthor}
\end{subfigure}
\caption{Prediction accuracy of FlowGN with different path length $l$ and in-out parameter $q$.}
\label{r-l path}
\end{figure}

In this section we explore how different flow path generation strategies affect the performance of FlowGN. To encourage flowability, we set the return parameter  $p=1000$ to reduce duplicate nodes in a flow path.  We first vary the in-out parameter $q\in \{0.1,0.5,1,2,4\}$ to interpolate between DFS and BFS strategies, and then vary path length $l$ to evaluate the effectiveness of FlowGN. The experimental results are shown in Figure \ref{r-l path}. We observe that FlowGN achieves its best performance when $q=0.1$ on all datasets, which indicates that DFS plays a more important role than BFS in FlowGN to learn node representations. In network analysis DFS is tied to the homophily hypothesis \citep{hoff2002latent}, so we can infer that nodes with similar structural roles in graphs will learn similar representations in \method{}. Moreover, we find that the optimal $l$ varies across datasets: $8$ in \emph{Citeseer}, $6$ in \emph{Pubmed}, and $5$ in \emph{Coauthor}.
Note that in Table \ref{table:citation-dataset} we computed the average shortest path length ($Avg. SP$), which are $9.33$, $6.34$, and $5.43$ for the three datasets respectively. $Avg. SP$ and the optimal $l$ seem to be positively correlated, and reasonable explanations are: 1) \method{} can learn the overall structure of graphs with $l=Avg. SP$. 2) Transmitting information through shortest paths is efficient and longer paths ($l > Avg. SP$) may cause redundant computation and lead to overfitting. On the other hand, GCNs implicitly take a ``flooding'' strategy to transmit information and the heavy design makes it difficult to explore long propagation ranges. We can conclude that \method{} is more flexible in propagation strategies.

\subsection{Effectiveness of Model Depth}  
\begin{figure}
\begin{subfigure}{0.3\textwidth}
\begin{tikzpicture}[scale=0.5]
\begin{axis}[
    xlabel={$\mathcal{K}$ },
    xmin=1, xmax=7,
    ymin=0.3, ymax=0.9,
    xtick={1,2,3,4,5,6,7},
    ytick={0,0.2,0.3,0.4,0.5,0.6,0.7,0.8,0.9 },
    legend pos=south west,
    ymajorgrids=true,
    grid style=dashed,
]
 \addplot[
    color=red,
    mark=*,
    line width=1pt,
    ]
    coordinates {
   (1,0.78663)(2,0.78219)(3,0.78219)(4,0.78339)(5,0.78960)(6,0.78460)(7,0.78410)
    };
    \addlegendentry{FlowGN}
    
   \addplot[
    color=violet,
    mark=diamond,
    line width=2pt,
    ]
    coordinates {
   (1, 0.4410)(2,0.6421)(3,0.6856)(4, 0.63636)(5, 0.6163)(6, 0.6042)(7, 0.5755)
    };
   \addlegendentry{GraphSAGE}
    
\addplot[
    color=cyan,
    mark=asterisk,
    line width=2pt,
    ]
    coordinates {
    (1,0.74900)(2,0.77600)(3,0.77000)(4,0.76000)(5,0.50300)(6,0.47100)(7,0.16900)
    };
   \addlegendentry{GCN}

\addplot[
    color=blue,
    mark=square,
     line width=1pt,
    ]
    coordinates {
  (1, 0.769)
(2, 0.769)
(3, 0.765)
(4, 0.76)
(5, 0.76)
(6, 0.7590000000000001)
(7, 0.757)
    };
     \addlegendentry{SGC}
\end{axis}
\end{tikzpicture}
\caption{Citeseer}
\end{subfigure}%
\quad\quad
\begin{subfigure}{0.3\textwidth}
\begin{tikzpicture}[scale=0.5]
\begin{axis}[
    xlabel={$\mathcal{K}$ },
    xmin=1, xmax=7,
    ymin=0.6, ymax=1.,
    xtick={1,2,3,4,5,6,7},
    ytick={0,0.2,0.3,0.4,0.5,0.6,0.7,0.8,0.9,1.0,0.65,0.75,0.85,0.95 },
    legend pos=south west,
    ymajorgrids=true,
    grid style=dashed,
]
\addplot[
    color=red,
    mark=*,
     line width=1pt,
    ]
    coordinates {
  (1,0.88471)(2,0.89078)(3,0.89320)(4,0.89320)(5,0.89442)(6,0.89078)(7,0.89199)
    };
    \addlegendentry{FlowGN}
    
    \addplot[
    color=violet,
    mark=diamond,
    line width=2pt,
    ]
    coordinates {
   (1, 0.8662)(2,0.8880)(3,0.8797)(4,0.8656)(5,0.8551)(6,0.8329)(7,0.8252)

    };
   \addlegendentry{GraphSAGE}   
   
    \addplot[
    color=cyan,
    mark=asterisk,
    line width=2pt,
    ]
    coordinates {
    (1,0.79400)(2,0.86600)(3,0.86900)(4,0.86200)(5,0.85100)(6,0.84200)(7,0.73400)
    };
   \addlegendentry{GCN}

\addplot[
    color=blue,
    mark=square,
     line width=1pt,
    ]
    coordinates {
    (1, 0.848)
(2, 0.8289999999999998)
(3, 0.819)
(4, 0.81)
(5, 0.81)
(6, 0.808)
(7, 0.805)
    };
    \addlegendentry{SGC}
     
\end{axis}
\end{tikzpicture}
\caption{Pubmed}
\end{subfigure}%
\quad\quad
\begin{subfigure}{0.3\textwidth}
\begin{tikzpicture}[scale=0.5]
\begin{axis}[
    xlabel={$\mathcal{K}$ },
    xmin=1, xmax=7,
    ymin=0.6, ymax=1,
    xtick={1,2,3,4,5,6,7},
    ytick={0,0.2,0.3,0.4,0.5,0.6,0.7,0.8,0.9,1.0,0.65,0.75,0.85,0.95 },
    legend pos=south west,
    ymajorgrids=true,
    grid style=dashed,
]
  \addplot[
    color=red,
    mark=*,
    line width=1pt,
    ]
    coordinates {
    (1,0.94597)(2,0.94393)(3,0.94088)(4,0.94190)(5,0.94393)(6,0.94292)(7,0.94190)
    };
    \addlegendentry{FlowGN}
    
       \addplot[
    color=violet,
    mark=diamond,
    line width=2pt,
    ]
    coordinates {
    (1,0.9182)(2,0.9211)(3,0.9175)(4,0.9050)(5,0.8829)(6,0.8710)(7,0.8710)
 
    };
   \addlegendentry{GraphSAGE}
\addplot[
    color=cyan,
    mark=asterisk,
    line width=2pt,
    ]
    coordinates {(1,0.74100)(2, 0.8570)(3,0.9090)(4,0.8850)(5,0.8180)(6,0.7910)(7,0.6880)
    };
   \addlegendentry{GCN}

 \addplot[
    color=blue,
    mark=square,
    line width=1pt,
    ]
    coordinates {
   (1, 0.912)
(2, 0.894)
(3, 0.867)
(4, 0.857)
(5, 0.839)
(6, 0.822)
(7, 0.815)
    };
    \addlegendentry{SGC}
    
\end{axis}
\end{tikzpicture}
\caption{Coauthor}
\end{subfigure}%
\caption{Prediction accuracy of FlowGN with different layer number $\mathcal{K}$}
\label{modeldepth}
\end{figure}
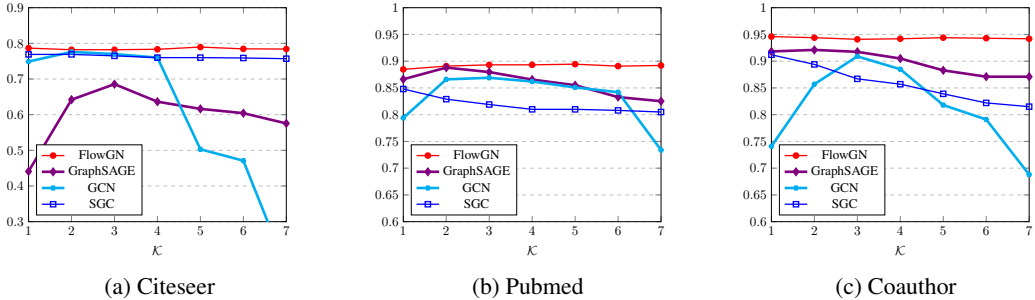

In the following experiments, we investigate the influence of model depth (number of layers) on the classification performance. We vary the layer number $K\in \{1,...,7\}$ to evaluate the effectiveness of FlowGN. The path generation follows the optimal strategies that we discussed in Section \ref{ana_flow_path}. Figure \ref{modeldepth} shows the experimental results on three datasets. We first find that \method{} is competitive with one-layer structure; it achieves better performance than other baselines with multiple-layer structures. This is because \method{} supports variable propagation range in one layer while other models are limited to one-hop propagation. Then we can see the performance of FlowGN first increases as we stack more layers on the model, reaching its peak at $K=5$. After that FlowGN tends to risk the overfitting problem and leads to a steady decrease in performance. In contrast, we observe that the performance of GCN drops dramatically when $K$ increases after $2$ (in \emph{Citeseer} and \emph{Pubmed}) or $3$ (in \emph{Coauthor}); GraphSAGE performs and SGC also leads to a deterioration when layer number increases. The empirical results reveal the coupling design leads to a performance deterioration. Moreover, the coupling design makes it hard to estimate the effects of model depth and propagation range separately. FlowGN achieves improvements with deeper structures via a decoupling design, showing a promising direction to address the shallow structure problem in GCNs.


\subsection{Time Complexity Analysis}
We select some state-of-the-art GCNs that support batch learning, and report their average training time per batch on four datasets in Table \ref{batch time}. For fair comparison, we set batch size as 256. Since one sample in \method{} (i.e., one flow path) has $l$ nodes, we use $256/l$ samples per batch for \method{}. We can see \method{} performs competitively; compared with other \emph{``Sink$\to$Source''}-mode GCNs (GCN and GraphSAGE), \method{} costs less time due to its explicit information propagation mechanism. We observe that FastGCN outperforms \method{} on every dataset. Note that FastGCN samples nodes independently across layer and ignores the dependence among nodes that are connected. FastGCN gains a high speedup with its sampling strategy but sacrifices its expressive capacity, which can be revealed from the prediction performance in Table \ref{table:baselines}.


 \begin{table}[ht]
 \small
\centering
\begin{tabular}{c|c|c|c|c}
\toprule
& Cora & Citeseer & Pubmed & Coauthor\\
\midrule
GCN (batched)            &0.0282 & 0.0868  &0.0746 & 1.3286\\
GraphSAGE-GCN  &0.8493 &1.8145 &0.3565 & 3.2760\\
FastGCN                  &0.0130 &0.0237 &0.0059 & 0.0421\\
\method{}                & 0.0185 &0.0423 &0.0414 & 0.0601\\
\bottomrule
\end{tabular}
\caption{Comparison of per-batch training time (in seconds).}
\label{batch time}
\end{table}

\section{Conclusion}
In this paper, we study the information propagation mechanism of GCNs in a \emph{``Source$\to$Sink''} mode. We introduce a new concept, ``information flow path", that explicitly defines where information originates and how it diffuses. We propose a framework named FlowGN that supports flexible propagation strategies and enables deep structures. Empirical results on real-world datasets show that our method outperform alternatives in prediction accuracy with less time complexity. In the future we will mainly extend \method{} in two directions. One direction is to design adaptive random walkers to generate flow paths that can transmit information more efficiently. The other direction is to design proper information propagation mechanism for representation learning in heterogeneous graphs.


\bibliographystyle{plainnat}
\bibliography{graph19-0}

\end{document}